\documentclass[11pt]{article}

\usepackage[margin=1in]{geometry}
\usepackage{moreverb,url}
\usepackage{makecell}
\usepackage{amsmath} 
\usepackage{mathptmx}
\usepackage{dsfont}
\usepackage{tcolorbox}
\usepackage{listings}
\usepackage{graphicx, epsfig, psfrag, svg}
\usepackage{color}
\usepackage{wrapfig, booktabs, float, longtable, lscape}
\usepackage{caption, subcaption}
\usepackage{tikz}
\usetikzlibrary{shapes,arrows}
\usepackage{multirow, makecell, array, diagbox}

\usepackage{dirtytalk}
\usepackage{enumitem}
\usepackage{lipsum}
\usepackage{fancyhdr}
\usepackage{titlesec}
\usepackage{pdfpages}
\usepackage[numbers,round,sort&compress]{natbib}
\usepackage[pagewise]{lineno}
\usepackage{nomencl}
\usepackage[colorlinks,bookmarksopen,bookmarksnumbered,citecolor=red,urlcolor=red]{hyperref}

\newcommand\BibTeX{{\rmfamily B\kern-.05em \textsc{i\kern-.025em b}\kern-.08em
T\kern-.1667em\lower.7ex\hbox{E}\kern-.125emX}}

\title{LLM-based Realistic Safety-Critical Driving Video Generation}

\author{%
Yongjie Fu$^{1}$, Ruijian Zha$^{2}$, Pei Tian$^{2}$, and Xuan Di$^{1}$\\[0.75ex]
\normalsize
$^{1}$Department of Civil Engineering and Engineering Mechanics, Columbia University, NY\\
\normalsize
$^{2}$Department of Computer Science, Columbia University, NY\\[0.5ex]
\small
Emails: \{yf2578, rz2689, pt2632, sharon.di\}@columbia.edu.\\
\small
Corresponding author: Xuan Di (\texttt{sharon.di@columbia.edu}).
}

\date{}

\begin{document}

\maketitle
\begin{center}
\small
\textit{Accepted for publication in Transportation Research Record.}
\end{center}
\vspace{0.75em}

\begin{abstract}
\noindent
Designing diverse and safety-critical driving scenarios is essential for evaluating autonomous driving systems. In this paper, we propose a novel framework that leverages Large Language Models (LLMs) for few-shot code generation to automatically synthesize driving scenarios within the CARLA simulator, which has flexibility in scenario scripting, efficient code-based control of traffic participants, and enforcement of realistic physical dynamics. Given a few example prompts and code samples, the LLM generates safety-critical scenario scripts that specify the behavior and placement of traffic participants, with a particular focus on collision events. To bridge the gap between simulation and real-world appearance, we integrate a video generation pipeline using Cosmos-Transfer1, which converts rendered scenes into realistic driving videos. Our approach enables controllable scenario generation and facilitates the creation of rare but critical edge cases, such as pedestrian crossings under occlusion or sudden vehicle cut-ins. Comprehensive quantitative evaluations across multiple environments demonstrate a favorable balance between visual realism, perceptual quality, and structural consistency in the generated videos. Experimental results demonstrate the effectiveness of our method in generating a wide range of realistic, diverse, and safety-critical scenarios, offering a promising tool for simulation-based testing of autonomous vehicles.
\end{abstract}


\vspace{1em}

\section{Introduction}
The rapid evolution of large language models (LLMs) and vision language models (VLMs) \cite{fu2024drivegenvlm, zha2025slider} has transformed numerous aspects of artificial intelligence, significantly impacting the autonomous vehicles (AVs) domain. Currently, LLMs are utilized across several key AV functionalities. For instance, they facilitate natural language interactions between vehicles and passengers, enhancing the in-car user experience \cite{cui2023drive}. LLMs have also been employed to generate diverse and realistic driving scenarios for robust AV testing and validation \cite{tan2023language}. Furthermore, LLMs are leveraged in interpreting complex sensor data and decision-making processes, improving the transparency and explainability of autonomous systems \cite{chen2023interpretable}. Additionally, recent research has integrated LLMs into traffic behavior prediction models, significantly improving accuracy in dynamic driving environments \cite{munir2024exploring}.

Generating safety-critical scenarios \cite{li2024safeaug} is essential for autonomous driving as it directly contributes to the robustness and reliability of AVs. Although typical driving conditions are often predictable, rare and hazardous scenarios pose significant safety challenges, potentially leading to severe accidents if not anticipated and managed effectively \cite{koopman2018autonomous}. These critical scenarios, such as sudden pedestrian crossings, unexpected lane changes, or adverse weather conditions, are infrequent in real-world driving data, making them difficult to thoroughly evaluate through traditional road testing alone. By synthetically generating such scenarios, developers can systematically test and validate the vehicle's perception, planning, and control algorithms under extreme conditions, ensuring the AVs system's resilience against uncommon yet high-risk events \cite{riedmaier2020survey}. Ultimately, the proactive identification and mitigation of safety-critical scenarios can significantly reduce accident rates and accelerate the deployment of trustworthy autonomous driving technologies.

Recent advancements have showcased the potential of LLMs to assist in simulation-oriented code generation, enabling rapid prototyping of complex tasks in robotics and autonomous driving. For instance, LangProp introduces an iterative feedback mechanism to refine LLM-generated code for autonomous driving scenarios in CARLA, enhancing safety and diversity through simulation-based evaluations \cite{langprop}. Similarly, GenSim leverages LLMs to synthesize robotic manipulation tasks and corresponding expert trajectories, demonstrating strong generalization in unseen environments \cite{gensim}. ChatScene further bridges natural language and simulation by translating high-level prompts into domain-specific scenario scripts, facilitating the creation of safety-critical events \cite{chatscene}. These efforts collectively highlight the growing capability of LLMs to autonomously generate, optimize, and validate code for realistic simulation environments, laying the groundwork for more efficient and scalable autonomous system development.

Recent advances in diffusion-based video generation have demonstrated remarkable capabilities in synthesizing photorealistic and temporally coherent videos from high-level inputs such as text or keyframes. Models like Video Diffusion Models \cite{ho2022videodiffusion}, Make-A-Video \cite{singer2022makeavideo}, and Phenaki \cite{bandarkar2023phenaki} illustrate the potential of leveraging diffusion processes to generate diverse and controllable video content. These approaches have significantly improved visual fidelity and semantic alignment, making them increasingly suitable for simulation and content creation tasks. However, their application to safety-critical domains like autonomous driving remains limited, particularly in the context of integrating structured simulation logic or scenario control.

To address this gap, we propose a novel framework that couples LLMs with Cosmos-Transfer1, a diffusion-based video generation model tailored for driving environments. While existing efforts primarily focus on generating code or simulated scenes in platforms such as CARLA, our method translates LLM-generated structured driving scenarios into realistic videos. By doing so, we enable high-fidelity visualization of rare and hazardous situations, such as vehicle-cyclist collision or vehicle-vehicle collision. This integration enhances the realism and utility of synthetic datasets, supporting perception and planning modules in autonomous vehicles through photorealistic, safety-critical training samples.

The main contributions of our work are summarized as follows:
\begin{itemize}
    \item \textbf{LLM-Driven Scenario Generation:} We propose a few-shot prompting approach using LLMs to automatically synthesize code for generating diverse and safety-critical driving scenarios, particularly collision scenarios, within the CARLA simulator
    
    \item \textbf{Controllable Realistic Video Synthesis:} We utilize Cosmos-Transfer1 to transform simulated outputs into realistic driving videos, enabling visual fidelity while preserving control over scene semantics.
    
    \item \textbf{End-to-End Scenario-to-Video Pipeline for AV Testing:} We develop a controllable, end-to-end pipeline that bridges simulation and video generation, facilitating the creation and analysis of edge-case scenarios critical for autonomous vehicle evaluation.

    \item \textbf{Comprehensive Quantitative Evaluation:} We provide an in-depth quantitative assessment of video generation quality across multiple simulated environments using a suite of established metrics, offering valuable benchmarks and insights for future research in simulation-to-realistic video synthesis.

\end{itemize}

The remainder of this paper is organized as follows. Sec.~\ref{sec-related} reviews related work on safety-critical driving scenario generation and realistic driving video synthesis. Sec.~\ref{sec-methodology} presents the proposed framework, including the LLM-based few-shot scenario generation in CARLA and the diffusion-based video synthesis using Cosmos-Transfer1. Sec.~\ref{sec-experiments} describes the experimental setup and provides both qualitative and quantitative evaluations of the generated scenarios and videos, including scenario validity, visual realism, and physics-consistency analyses. Finally, Sec.~\ref{sec-conclusion} concludes the paper and discusses directions for future work.
\section{RELATED WORK}
\label{sec-related}

\subsection{Safety Critical Driving Scenario Generation}
The generation of safety-critical scenarios for AVs generally falls into three main categories:

\textbf{Data-driven generation:} This approach extracts patterns and edge cases from real-world driving data to construct realistic scenarios. Recent methods employ generative models trained on large-scale datasets to produce novel yet statistically plausible traffic situations \cite{tan2021scenegen}. While providing naturalistic scenarios, these approaches may struggle to generate sufficiently rare edge cases that are underrepresented in datasets \cite{fu2024gendds}.

\textbf{Adversarial generation:} This methodology actively seeks to identify vulnerabilities in AV systems by generating scenarios that maximize failure rates. Techniques range from gradient-based optimization \cite{wang2021advsim} to reinforcement learning approaches that train adversarial agents to create challenging situations \cite{sharif2021adversarial}. These methods efficiently uncover edge cases but may produce physically implausible scenarios requiring post-processing.

\textbf{Knowledge-based generation:} This category utilizes domain expertise to craft scenarios based on known risk factors and safety requirements. Recent work integrates ontologies and formal safety specifications with generative techniques to ensure both diversity and criticality \cite{bagschik2018ontology}. These approaches benefit from human expertise but can be limited by the expressiveness of their knowledge representation.

\subsection{Driving Video Generation}
Recent advancements in video generation have significantly impacted autonomous driving research. Panacea \cite{wen2024panacea} is a panoramic controllable driving video generation framework designed for autonomous driving applications, enabling environment adaptations (e.g., viewpoint and scene layout changes) while maintaining semantic consistency. Similarly, DriveDreamer-2 leveraged large language models to enhance world models for diverse driving video generation, demonstrating the growing integration of language understanding with visual synthesis \cite{zhao2025drivedreamer}.

Cosmos-Transfer1 builds upon these foundations by offering adaptive multimodal control with superior fidelity \cite{alhaija2025cosmos}. Unlike previous approaches that often require extensive fine-tuning, Cosmos-Transfer1's pre-trained capabilities allow direct application to driving scenarios, streamlining the development process. The model's ability to process multiple control modalities simultaneously represents a significant advancement over earlier methods that primarily relied on single modality conditioning.

Other notable works include Stag-1, which focuses on realistic 4D driving simulation \cite{wang2024stag}, and various GAN-based approaches that prioritize temporal consistency. However, Cosmos-Transfer1's diffusion-based architecture provides advantages in terms of both quality and controllability, making it particularly valuable for safety-critical scenario generation where precise control over environmental conditions is essential.

\section{METHODOLOGY}
\label{sec-methodology}

\begin{figure*}[h]
    \centering
    \includegraphics[width=0.7\linewidth]{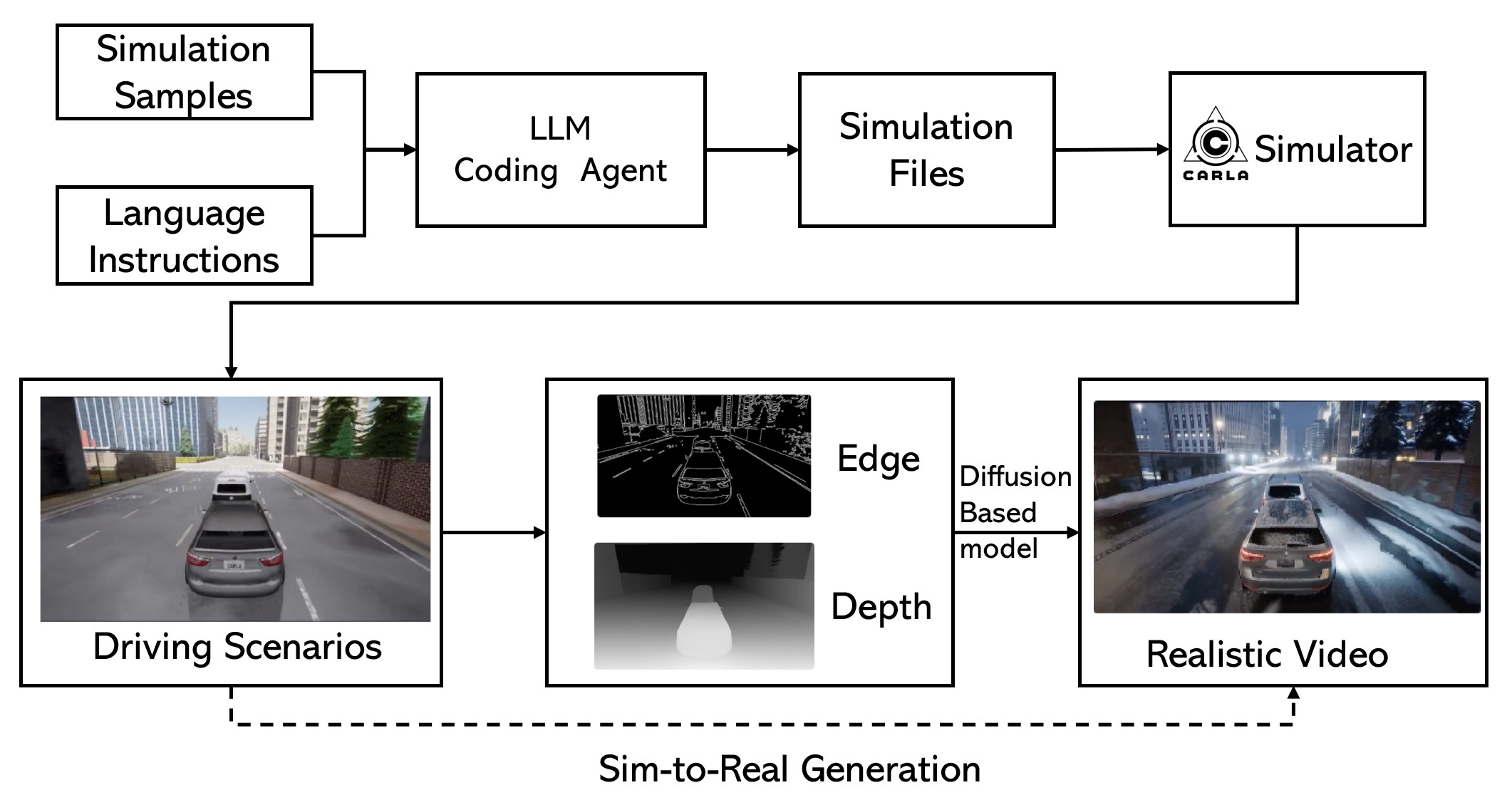}
    \caption{Framework for LLM-driven scenario generation and Cosmos-Transfer1 video synthesis. Our pipeline consists of two main stages: (1) LLM-based scenario generation in CARLA using few-shot prompting, which produces traffic simulations with safety-critical events; and (2) Realistic video synthesis using Cosmos-Transfer1, which transforms the simulated outputs into photorealistic driving videos with diverse environmental conditions.}
    \label{fig:framework}
\end{figure*}

Our framework, illustrated in Fig.~\ref{fig:framework}, integrates LLM-driven scenario generation with photorealistic video synthesis to support autonomous vehicle testing in diverse and safety-critical environments. It comprises two main components: LLM-based scenario generation (top) and realistic video generation (bottom). In the first stage, we leverage LLMs with few-shot prompting to generate diverse traffic scenarios within the CARLA simulator. These scenarios are designed to include rare and safety-critical events, such as pedestrian crossings, sudden stops, or multi-agent interactions. In the second stage, we employ Cosmos-Transfer1, a state-of-the-art video synthesis model, to transform the simulated outputs into realistic driving videos. This process enhances the visual fidelity of the generated scenes by simulating various environmental conditions such as weather, lighting, and road textures. Together, these components enable the scalable creation of high-quality, diverse, and challenging datasets for testing and evaluating autonomous driving systems. We introduce each component in detail in the following sections.

\subsection{LLM-based Scenario Generation}

\subsubsection{Few-shot Prompting for Code Generation}
We utilize Scenic \cite{fremont2019scenic}, a domain-specific probabilistic programming language, to script scenes within the CARLA simulator. To support the generation of high-quality scenic scripts for scenario simulation, we adopt a few-shot learning approach using domain-specific LLMs, namely OpenAI's o4-mini ~\cite{openai_o4mini_2025} and Alibaba's Qwen2.5-Coder-32B-Instruct~\cite{hui2024qwen2}. These models are pre-trained on a mixture of natural language and code-related corpora, making them particularly suitable for generation tasks involving structured scripting logic. Few-shot learning is employed to condition the models on a small set of example scripts, which define the desired formatting, semantic structure, and narrative logic required for simulation environments. This method leverages the LLMs' extensive pretraining and instruction-following capabilities~\cite{brown2020language}. As a result, the models are able to generate contextually coherent and syntactically valid scripts that align with domain-specific constraints, allowing scalable content generation for complex and evolving scenario simulations. 

\begin{table*}[h]
\centering
\begin{tabular}{p{0.95\linewidth}}
\toprule
\textbf{Prompt Template for Few-Shot Script Generation} \\
\midrule

You are a helpful assistant. Please review the backbone and syntax of the following Scenic scripts for general driving scenarios. Based on these examples, try to generate a script for a collision scenario (e.g., pedestrian collision, T-bone collision, rear-end collision). \\[0.5em]

\textbf{Examples of Scenic scripts for driving scenarios:} \\[0.3em]

\textbf{\{Scenic script example\}} \\[0.3em]

{\ttfamily
\begin{minipage}[t]{0.93\linewidth}
try: \\
\hspace*{1.5em}do FollowLaneBehavior(EGO\_SPEED, laneToFollow=current\_lane) \\[0.3em]

interrupt when (distance to obstacle) < DIST\_THRESHOLD and not laneChangeCompleted: \\
\hspace*{1.5em}if ego can see oncomingCar: \\
\hspace*{3em}take SetBrakeAction(0.8) \\
\hspace*{1.5em}elif (distance to oncomingCar) > YIELD\_THRESHOLD: \\
\hspace*{3em}do LaneChangeBehavior(path, is\_oppositeTraffic=True, target\_speed=EGO\_SPEED) \\
\hspace*{3em}do FollowLaneBehavior(EGO\_SPEED, is\_oppositeTraffic=True) \\
\hspace*{3em}until (distance to obstacle) > BYPASS\_DIST \\
\hspace*{3em}laneChangeCompleted = True \\
\hspace*{1.5em}else: \\
\hspace*{3em}wait
\end{minipage}
} \\[0.5em]

\textbf{\{Scenic script example\}} \\[0.5em]

The generated scenario should follow the behavior logic below: \\[0.5em]

Your generated Scenic script: \\
\bottomrule
\end{tabular}
\caption{Prompt template used for few-shot learning to generate collision scenarios using Scenic scripts, including explicit behavior logic constraints.}
\label{tab:prompt_template}
\end{table*}

Chatscene \cite{chatscene} adopts an indirect approach that leverages LLMs to first curate a retrieval database of Scenic code snippets, encompassing fundamental elements of driving scenarios. While Chatscene mainly generates near-miss scenarios, we build upon these by extending and modifying them to create safety-critical scenarios where actual collisions occur. We leverage the capabilities of LLMs to translate scenario descriptions into scenic code. Through few-shot prompting, we provide the LLM with example pairs of scenario descriptions and their corresponding code implementations, enabling it to learn the mapping between natural language specifications and code patterns without extensive fine-tuning.

As illustrated in Tab.~\ref{tab:prompt_template}, when prompted with a request such as “try to generate a script for a collision scenario (e.g., pedestrian collision, T-bone collision, rear-end collision),” the LLM produces a complete Scenic script that specifies the precise positioning of the ego vehicle, surrounding parked vehicles, and pedestrians, along with temporal triggers for crossing events. This method enables rapid adaptation to various scenario types and simulation environments. By adjusting trigger thresholds and relative spatial configurations in the script, the LLM can effectively transform near-miss scenarios into actual collision events.


\subsubsection{Safety-Critical Scenario Types}
Our framework focuses on generating diverse safety-critical scenarios, including:
\begin{itemize}
    \item Sudden pedestrian crossings under occlusion
    \item Vehicle cut-ins with minimal warning
    \item Intersection conflicts with obstructed visibility
    \item Lane changes during adverse weather conditions
\end{itemize}
\begin{figure*}[h]
    \centering
    \begin{subfigure}[b]{0.45\textwidth}
        \includegraphics[width=\textwidth]{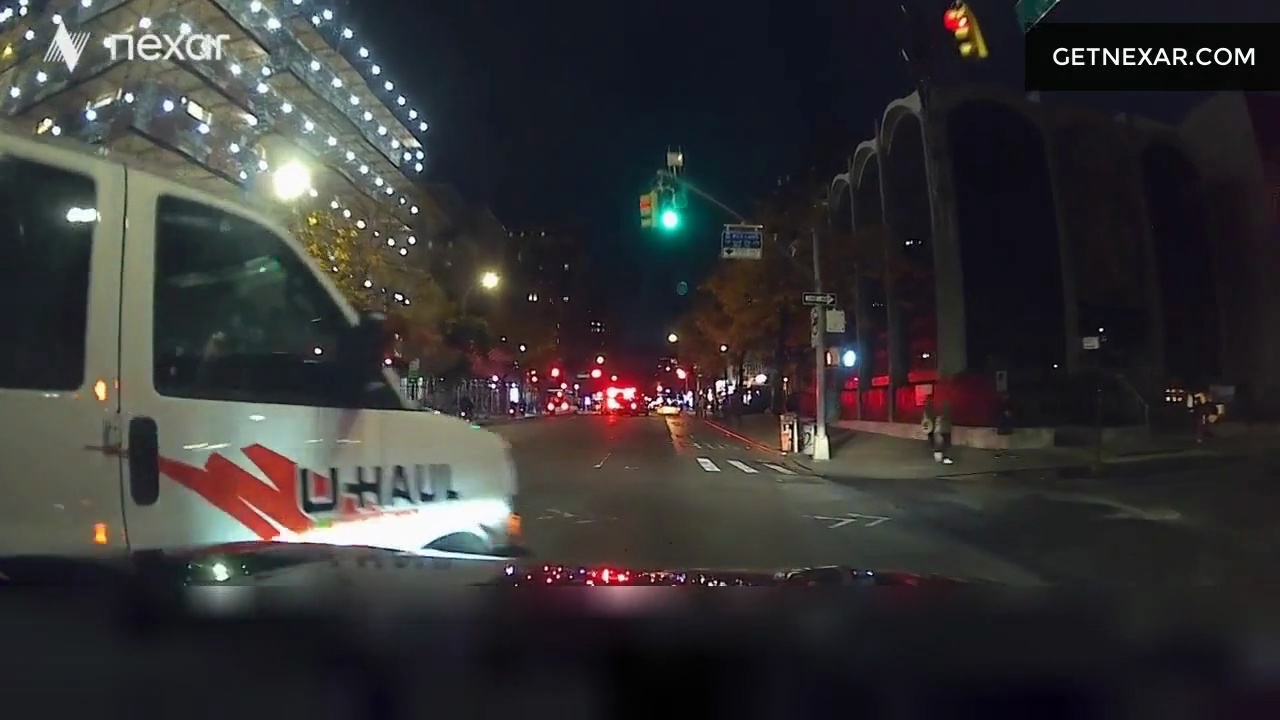}
        \caption{Original Video}
        \label{fig:original}
    \end{subfigure}
    \hfill
    \begin{subfigure}[b]{0.45\textwidth}
        \includegraphics[width=\textwidth]{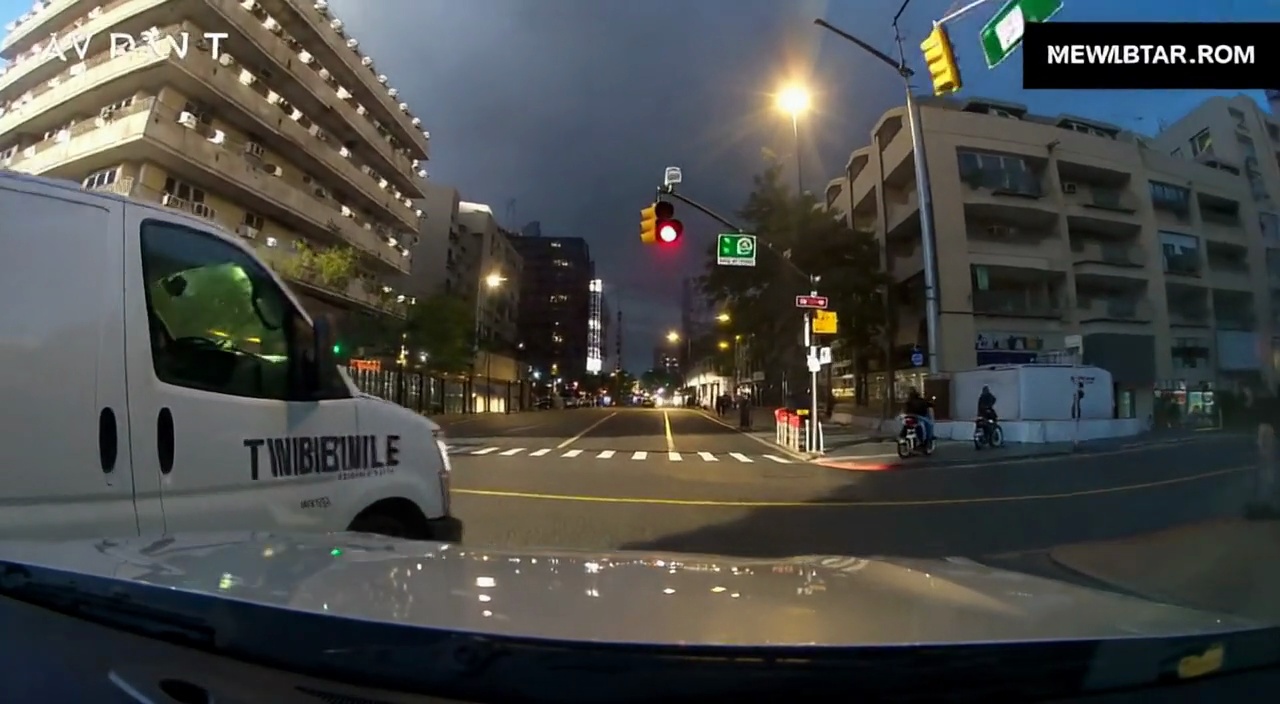}
        \caption{Night to Daytime}
        \label{fig:daytime}
    \end{subfigure}

    \vspace{0.5em} 

    \begin{subfigure}[b]{0.45\textwidth}
        \includegraphics[width=\textwidth]{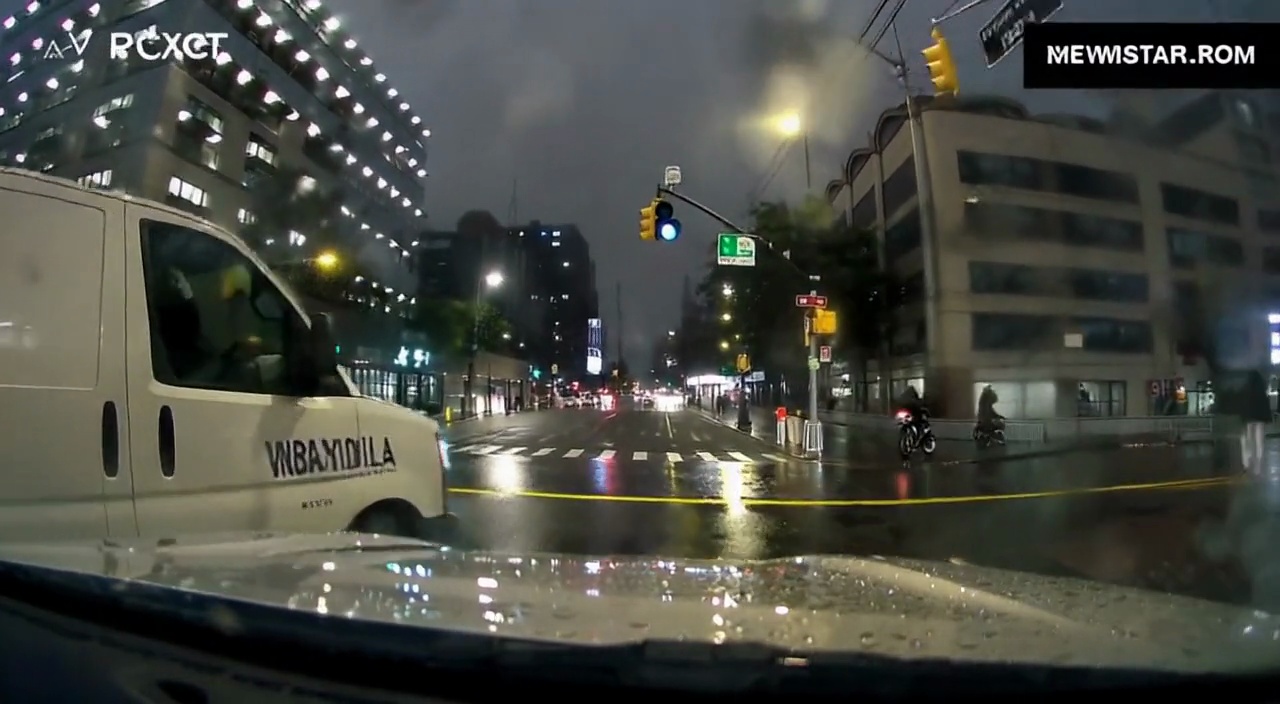}
        \caption{Rainy conditions}
        \label{fig:rainy}
    \end{subfigure}
    \hfill
    \begin{subfigure}[b]{0.45\textwidth}
        \includegraphics[width=\textwidth]{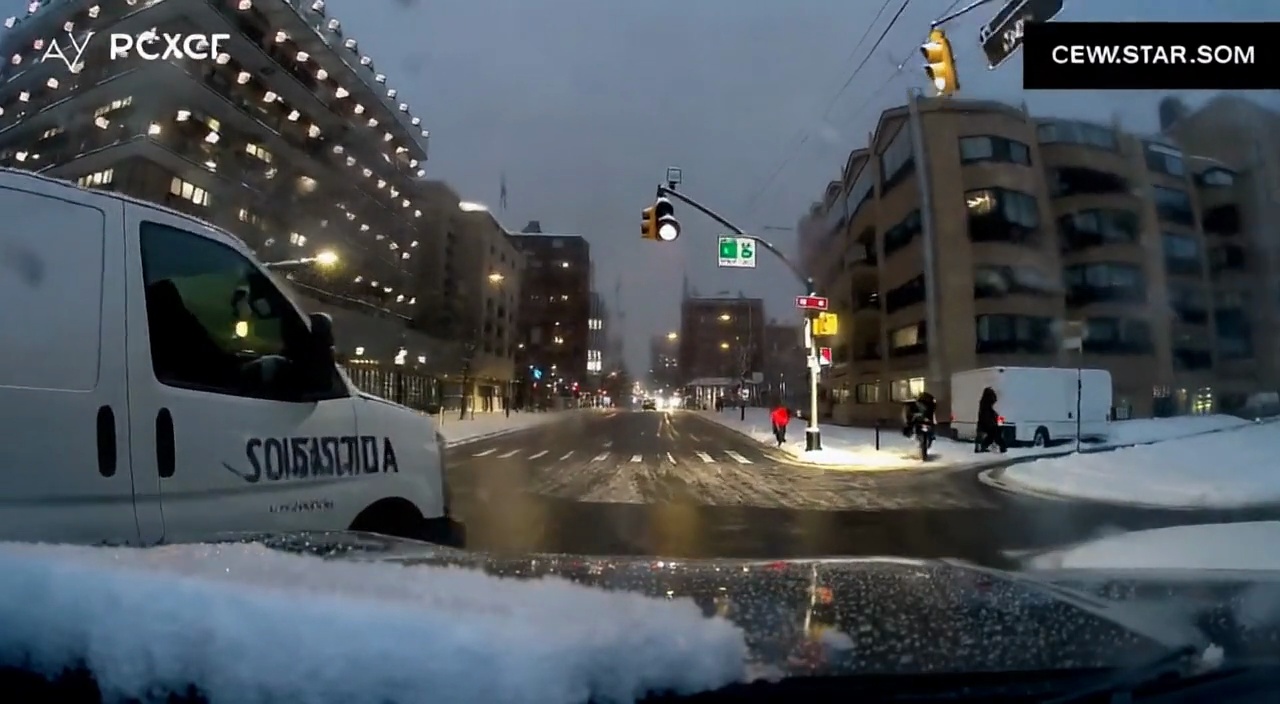}
        \caption{Snowy conditions}
        \label{fig:snowy}
    \end{subfigure}
    
    \caption{Comparison of original video (a) and Cosmos-Transfer1 enhanced environmental variations (b-d). The model successfully transforms the same safety-critical scenario (vehicle collision) into different environmental conditions while maintaining semantic consistency and adding realistic environmental effects.}
    \label{fig:variations}
\end{figure*}

The LLM's code generation capabilities allow for precise control over the timing, positioning, and behaviors of all traffic participants, creating reproducible scenarios that target specific edge cases. Furthermore, the natural language interface enables rapid iteration and customization without requiring expert programming knowledge.

\subsection{Cosmos-Transfer1 for Realistic Video Generation}
To bridge the gap between simulation and reality, we employ Cosmos-Transfer1~\cite{alhaija2025cosmos}, a diffusion-based conditional world model developed for multimodal controllable world generation.

\subsubsection{Architecture and Control Mechanisms} Cosmos-Transfer1 leverages ControlNet technology to generate high-fidelity videos conditioned on various spatial control inputs. The model operates in a latent space using a diffusion transformer, where different control branches process spatiotemporal control maps. Let $z_0$ denote the initial latent representation sampled from a Gaussian distribution, and $z_T$ denote the final denoised output. The diffusion process iteratively refines $z_t$ through a conditional denoising function $D_{\theta}$ guided by control inputs $C$: \begin{equation} z_{t-1} = D_{\theta}(z_t, C, t), \end{equation} where $t$ is the diffusion timestep, and $C$ includes the spatial and textual control signals. Control branches inject these signals through interleaved self-attention, cross-attention, and feedforward layers to ensure alignment between the generated content and the input conditions.

\subsubsection{Multi-modal Input Processing} Our implementation extracts control modalities from CARLA, specifically segmentation maps and depth information, which serve as structural guidance for the video generation process. The control modalities are combined into a unified control input $C$ via adaptive weighting: \begin{equation} C = w_{\text{seg}} \times C_{\text{seg}} + w_{\text{depth}} \times C_{\text{depth}}, \end{equation} where $C_{\text{seg}}$ and $C_{\text{depth}}$ denote the segmentation and depth maps, and $w_{\text{seg}}$, $w_{\text{depth}}$ are their respective weights. Additionally, Cosmos-Transfer1 incorporates text prompts $p$ that specify environmental attributes such as time of day, weather, and lighting. The overall conditioning can thus be represented as $C$ and $p$. 

\subsubsection{Adaptive Weighting for Visual Consistency} The adaptive weighting mechanism balances the contributions of different modalities to achieve both structural consistency and visual diversity. The weights $w_{\text{seg}}$ and $w_{\text{depth}}$ are adjusted based on the scenario requirements to ensure that critical semantic features (e.g., traffic participant positions) are preserved, while stylistic variations (e.g., road appearance, lighting) are realistically enhanced. This design enables Cosmos-Transfer1 to maintain the safety-critical aspects of simulated scenarios while significantly improving their visual fidelity.

Pre-trained on approximately 20 million hours of video data, Cosmos-Transfer1 can be applied directly without additional fine-tuning. This capability makes it an ideal tool for enhancing synthetic driving scenarios with realistic visual elements, thereby supporting robust simulation-to-reality transfer for autonomous vehicle testing.

\section{EXPERIMENTS}
\label{sec-experiments}

\begin{table*}[htbp]
    \centering
    \caption{LLMs Generated Scenarios in CARLA.}
    \renewcommand{\arraystretch}{1.5}
    \begin{tabular}{|p{2.3cm}|c|c|c|}
        \hline
        \textbf{Description} & \textbf{Scenario A} & \textbf{Scenario B} & \textbf{Scenario C} \\
        \hline
        \makecell{Vehicle-cyclist\\ Collision} & \includegraphics[width=0.25\textwidth]{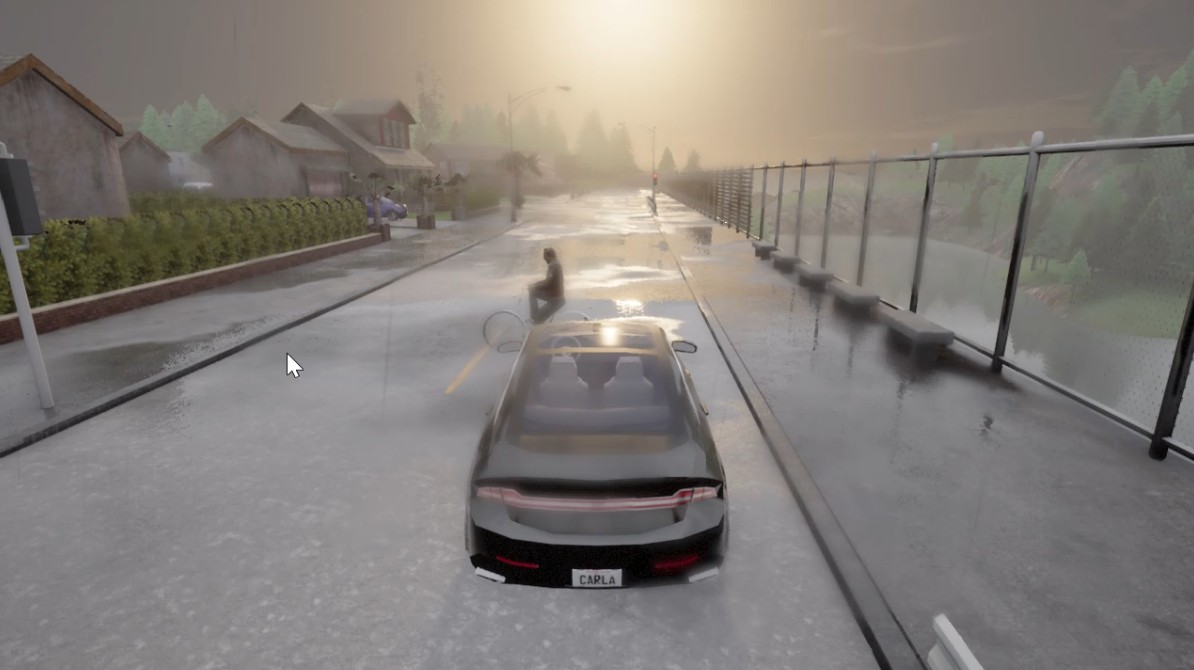} & \includegraphics[width=0.25\textwidth]{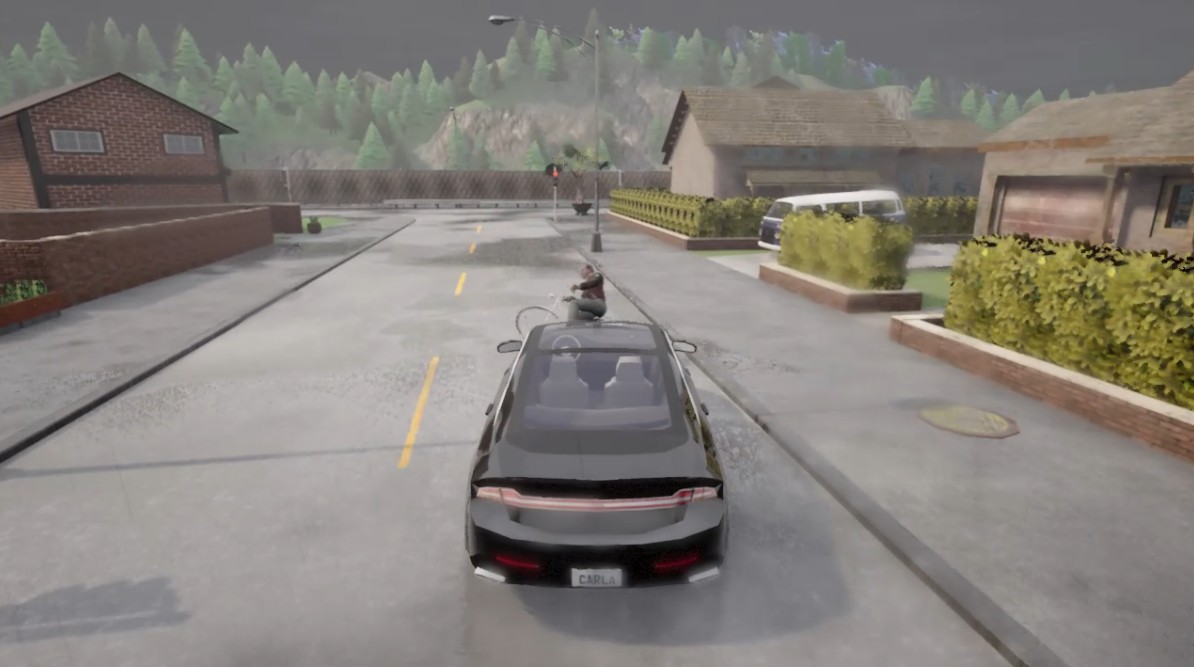} & \includegraphics[width=0.25\textwidth]{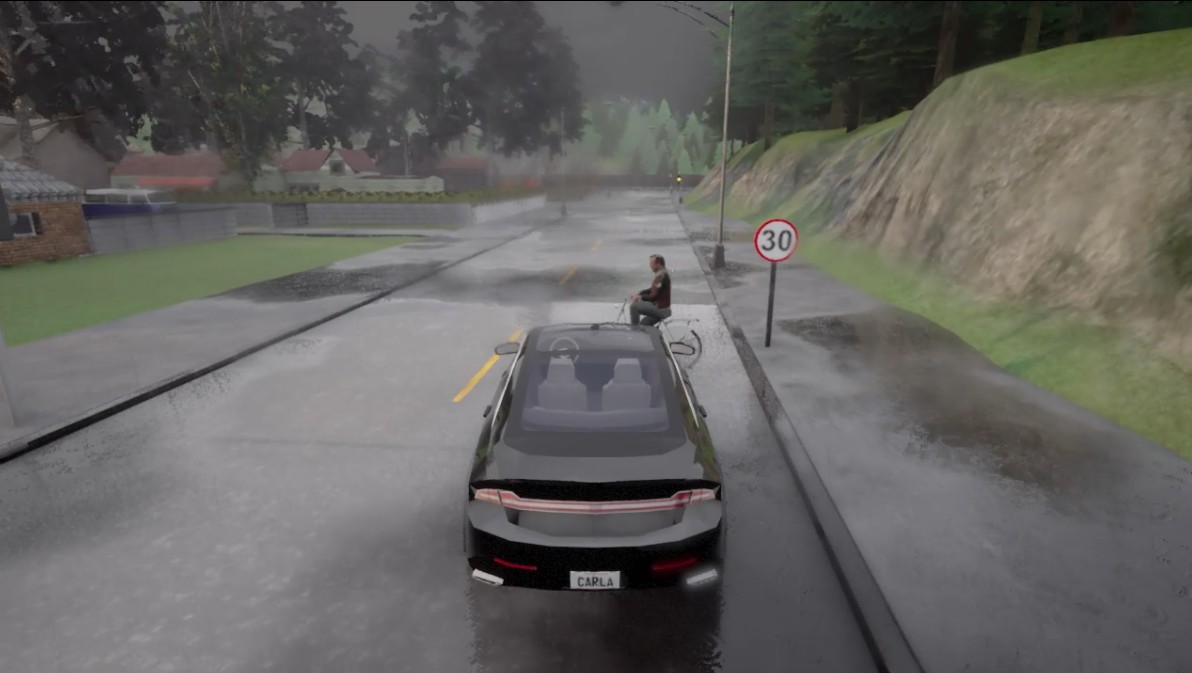} \\
        \hline
        \makecell{T-bone\\ Collision} & \includegraphics[width=0.25\textwidth]{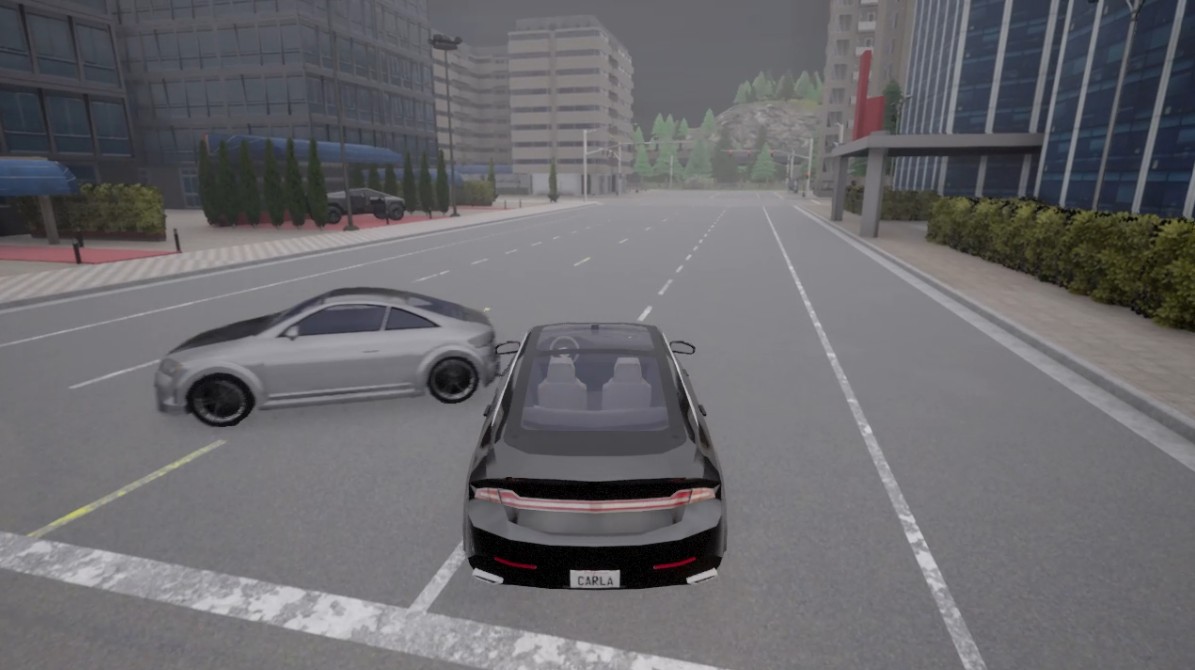} & \includegraphics[width=0.25\textwidth]{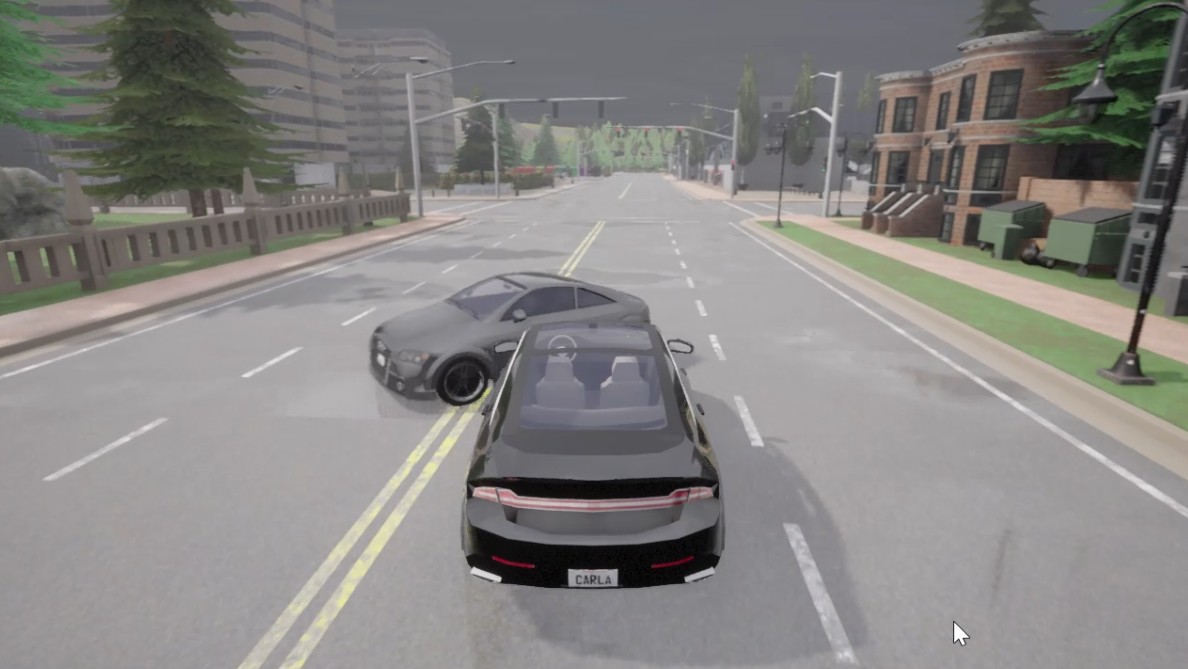} & \includegraphics[width=0.25\textwidth]{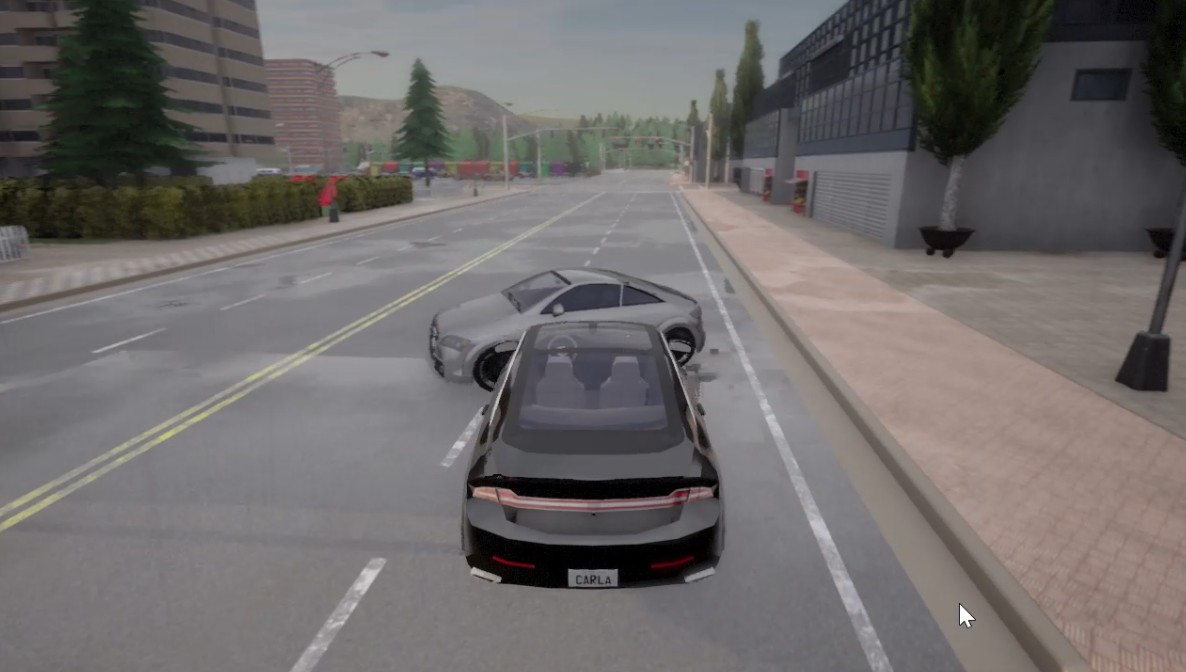} \\
        \hline
        \makecell{Rear-end\\ Collision} & \includegraphics[width=0.25\textwidth]{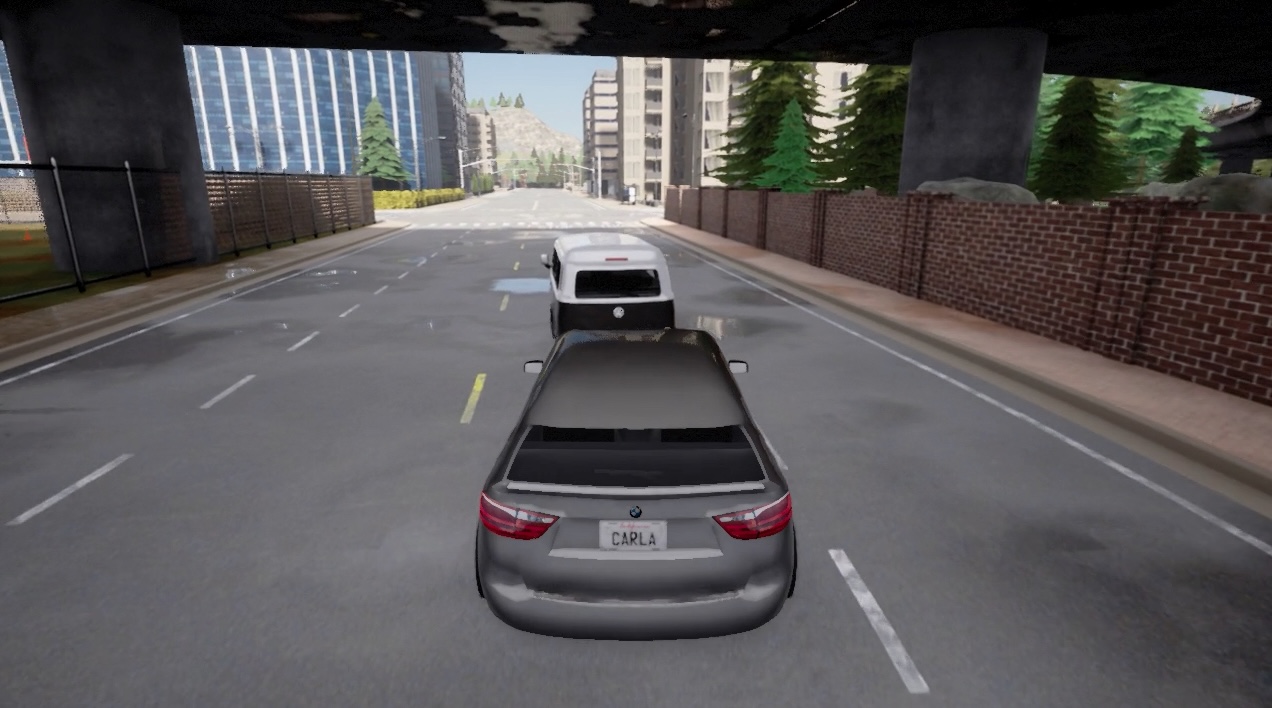} & \includegraphics[width=0.25\textwidth]{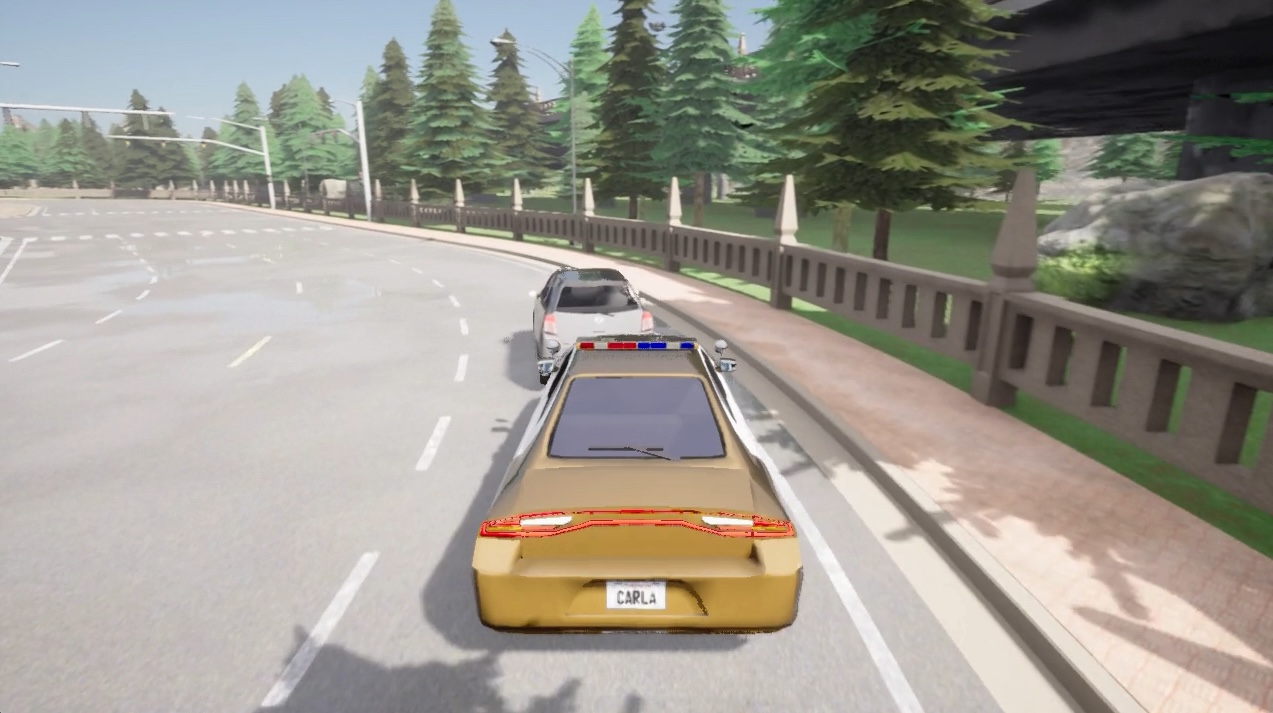} & \includegraphics[width=0.25\textwidth]{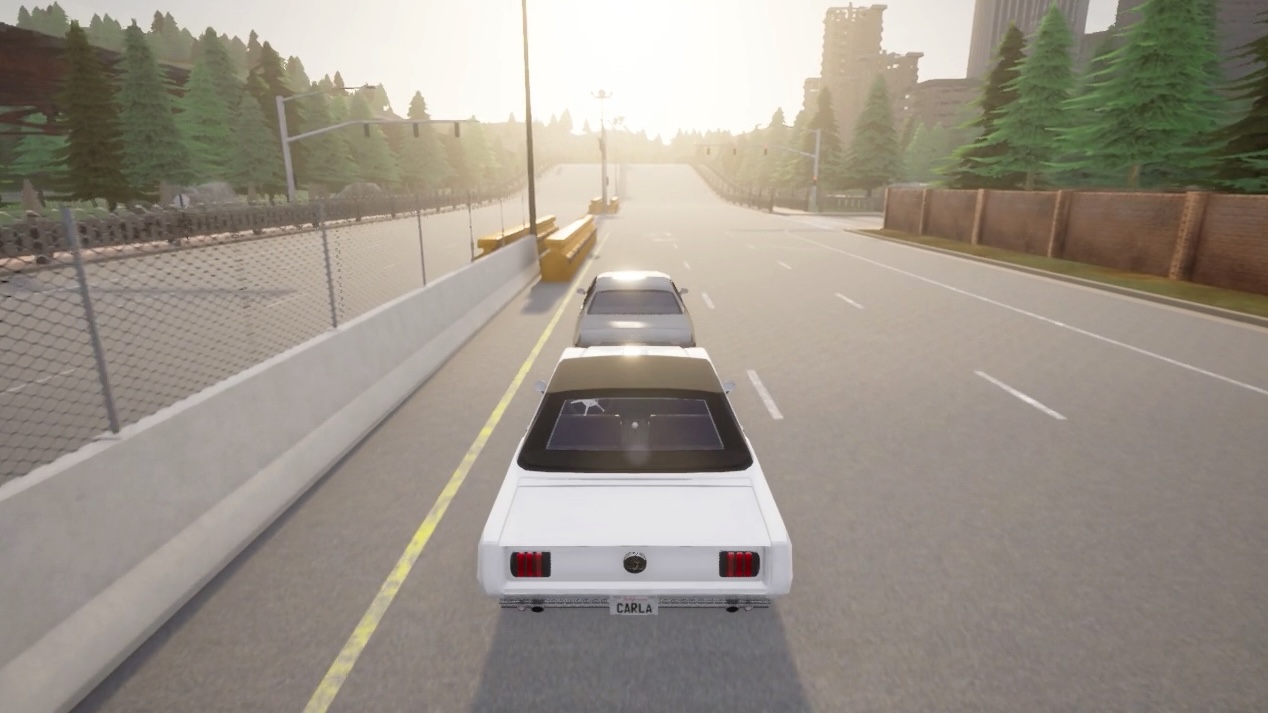} \\
        \hline
    \end{tabular}
    \label{tab:example_table}
\end{table*}

\subsection{Experimental Setup}

We evaluate our framework using scenarios generated in CARLA, which offers diverse urban and rural environments featuring various intersection types and road configurations. For each safety-critical scenario, we employ our LLM-based approach to generate $30$ distinct variations. The scenario generation leverages simulation examples from Chatscene \cite{chatscene}, formatted in Scenic, for few-shot learning.

Experiments are conducted in CARLA (v0.9.15), running on a desktop equipped with an RTX 4090 GPU and 7900X CPUs, to generate urban driving scenarios with complex traffic patterns. Edge and depth maps are generated by the preprocessor in Cosmos-Transfer1 from CARLA-rendered frames and used as control modalities. Additionally, text prompts such as ``sunny day,'' ``foggy evening,'' and ``rainy night'' guide the generation of diverse environmental conditions.

For inference, we employ the Cosmos-Transfer1-7B model on a single NVIDIA H100 GPU with 50 diffusion steps and a control strength of 0.8, balancing fidelity to input conditions with diversity in visual representation. As a baseline for video generation quality, we compare our approach against CogVideo, a state-of-the-art text-to-video generation model.

\subsection{Generation Performance}

\subsubsection{LLM-based Scenario Generation Performance}
Our LLM-based approach successfully generates diverse and complex traffic scenarios within CARLA. The few-shot prompting methodology proves effective in translating natural language descriptions into functional simulation code. The generated scenarios exhibit significant diversity in terms of traffic participant behaviors, timing, and spatial configurations, while maintaining the safety-critical characteristics specified in the prompts.

Tab.~\ref{tab:example_table} illustrates three types of generated collision scenes: vehicle-cyclist collisions, vehicle-to-vehicle T-bone collisions, and vehicle-to-vehicle rear-end collisions. Each collision type includes three distinct sample scenarios (Scenarios A, B, and C) set across diverse environments.

In addition to qualitative examples, we quantitatively evaluate the reliability of the proposed LLM-based scenario generation framework. Tab.~\ref{tab:success_rate} reports the success rate of generating valid, executable collision scenarios across the three collision categories. A generation is considered successful if the produced CARLA script executes without errors and results in the intended collision type as specified in the prompt. The results demonstrate consistently high success rates across different collision modes, indicating that the proposed few-shot prompting strategy is robust for safety-critical traffic scenario synthesis.

\begin{table*}[t]

\centering
\caption{Success rate of LLM-based generation for different collision scenario types in CARLA.}
\label{tab:success_rate}
\begin{tabular}{lccc}
\toprule
\textbf{Collision Type} & \textbf{Total Attempts} & \textbf{Successful Generations} & \textbf{Success Rate (\%)} \\
\midrule
Vehicle--Cyclist Collision        & 30 & 27 & 90.0 \\
Vehicle--Vehicle T-bone Collision & 30 & 26 & 86.7 \\
Vehicle--Vehicle Rear-end Collision & 30 & 28 & 93.3 \\
\midrule
\textbf{Overall} & 90 & 81 & 90.0 \\
\bottomrule
\end{tabular}

\end{table*}


\subsubsection{Video Generation Performance}
Cosmos-Transfer1 produces videos with enhanced visual fidelity, effectively capturing realistic weather and lighting variations while preserving the semantic structure of the original CARLA scenes.

\begin{figure*}[h]

    \centering
    \includegraphics[width=0.9\linewidth]{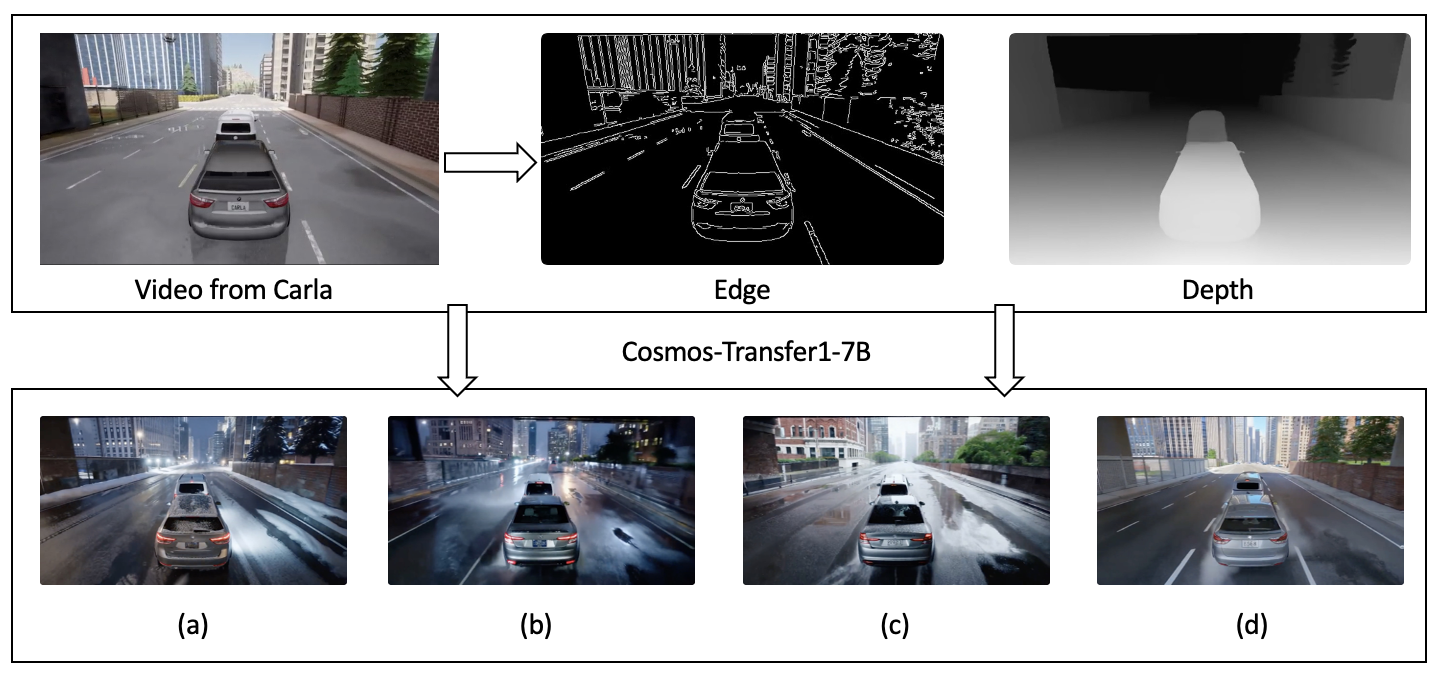}
    \caption{Realistic Video Synthesis from CARLA Simulations Using Cosmos-Transfer1}
    \label{fig:res-rear-end}

\end{figure*}

Fig.~\ref{fig:variations} showcases comparative results between original video renderings and Cosmos-Transfer1 enhanced videos across different environmental conditions. The qualitative results demonstrate Cosmos-Transfer1's ability to maintain semantic consistency while introducing realistic environmental variations. Notably, the model effectively renders complex lighting interactions in the daytime scenario, realistic water reflections and droplet effects in the rainy scenario, and appropriate snow accumulation patterns in the snowy scenario. These enhancements significantly improve visual realism without compromising the underlying scenario structure, validating our approach for safety-critical autonomous vehicle testing.

Fig.~\ref{fig:res-rear-end} showcases the results of our video generation pipeline. Starting from an original video generated in the CARLA simulator using LLM-based few-shot scenario synthesis, we extract edge maps and depth maps as structural inputs. These, along with text prompts specifying location and weather conditions, are provided to Cosmos-Transfer1-7B to produce realistic driving videos. As shown in the figure, the outputs generated by Cosmos-Transfer1 exhibit significantly enhanced visual realism compared to the original CARLA renderings. While the CARLA videos preserve the semantic structure, they often appear synthetic and lack fine visual details. In contrast, Cosmos-Transfer1 enriches the scenes with realistic textures, lighting variations, and environmental effects, resulting in photorealistic videos that are much closer to real-world driving footage. To further refine the output quality, we apply different adaptive weightings between the depth and edge control modalities for the four examples. Specifically, the weighting configurations are set as $(w_{\text{depth}}, w_{\text{edge}}) = (0.3, 0.4)$, $(0.2, 0.4)$, $(0.1, 0.4)$, and $(0.5, 0.5)$, respectively, where $w_{\text{depth}}$ and $w_{\text{edge}}$ denote the contribution of the depth and edge maps. This adaptive weighting ensures a flexible balance between spatial structure preservation and visual appearance enhancement across different scenarios.

\paragraph*{Detailed Scenario Descriptions.}
Fig.~\ref{fig:res-rear-end} depicts four examples of collision scenarios synthesized using 
Cosmos-Transfer1 under different environmental conditions.

\textbf{(a) Snowy night.} 
A silver sedan drives through snow-covered NYC streets with limited visibility and slippery 
pavement. Headlights illuminate dense snowfall as the vehicle struggles to maintain traction, 
ultimately rear-ending a white van beneath a dimly lit overpass. The environment includes 
snow-covered brick walls, softened building silhouettes, and dim streetlights reflecting off 
accumulated snow.

\textbf{(b) Rainy night.}
The sedan drives cautiously in heavy rainfall with low visibility and slick pavement, eventually 
rear-ending a van under the same overpass. The scene includes rain-soaked walls, reflective wet 
ground, fog-obscured lighting, and water spray from the sedan’s tires.

\textbf{(c) Rainy daytime.}
The sedan navigates wet NYC roads at moderate speed, approaching and rear-ending the van. 
The urban setting includes mist-covered buildings, glistening pavement, and greenery dripping 
with rain. Reflections from muted streetlights enhance realism.

\textbf{(d) Clear daytime.}
In clear daylight, the sedan again approaches and collides with the van. Brick walls, typical 
cityscape buildings, greenery, and road texture remain visible. This setting provides a clearer 
comparison of lighting and environmental variation across scenarios.

\subsection{Quantitive evaluation}
\subsubsection{Reference-based Video Quality Evaluation}

To quantitatively assess the quality of the generated driving videos, we evaluate multiple complementary aspects of visual and temporal quality using established metrics. Specifically, we assess temporal consistency, structural similarity, signal fidelity, and perceptual similarity using Fréchet Video Distance (FVD) \cite{unterthiner2019fvd}, Structural Similarity Index (SSIM) \cite{wang2004image}, Peak Signal-to-Noise Ratio (PSNR) \cite{hore2010image}, and Learned Perceptual Image Patch Similarity (LPIPS) \cite{zhang2018unreasonable}, respectively.

Fréchet Video Distance (FVD) measures the distributional distance between generated and reference video sequences, capturing both visual quality and temporal coherence. In simulation-to-real translation, a lower FVD indicates stronger temporal consistency, although excessively low values may also imply limited domain transformation. Therefore, in our setting, FVD is primarily interpreted as a measure of temporal stability rather than absolute similarity to the synthetic simulation domain.

Structural Similarity Index (SSIM) evaluates frame-level structural preservation by comparing luminance, contrast, and structural information. Due to the substantial appearance gap between CARLA-rendered videos and realistic outputs, SSIM values in the range of 0.3–0.6 are common in image generation and domain translation tasks, and relative improvements over baselines are generally more informative than absolute scores~\cite{isola2017image, wang2018video, yi2020survey}.

Peak Signal-to-Noise Ratio (PSNR) quantifies pixel-level fidelity between generated and reference frames. In generative and domain translation settings, PSNR values are typically lower than those reported in traditional image restoration tasks; values in the range of 14–20~dB are widely observed, with relative gains over baselines considered more meaningful than absolute magnitude~\cite{wang2018video, yi2020survey, gong2019dlow}.

Learned Perceptual Image Patch Similarity (LPIPS) measures perceptual similarity using deep neural network features that correlate well with human visual perception. In image and video generation tasks involving significant style or domain shifts, LPIPS values around 0.2–0.3 are commonly reported, and lower values indicate better perceptual alignment~\cite{zhang2018unreasonable, wang2018video, yi2020survey}.

Together, these metrics provide a comprehensive and complementary evaluation of temporal coherence, structural preservation, signal fidelity, and perceptual realism in the generated driving videos.

Tab.~\ref{tab:overall_performance} presents the overall performance of the video generation pipeline across multiple evaluation metrics. The FVD score of 1554.51 with a standard deviation of 472.76 indicates moderate temporal consistency in the generated videos, with some variation between different scenarios. The SSIM value of 0.4566 suggests fair preservation of structural information compared to the simulation input, which is expected given the significant domain gap. The PSNR value of 15.36~dB reflects acceptable signal fidelity, supporting that the generated content is reasonably close to the input in terms of pixel-wise similarity. Finally, the LPIPS score of 0.2794 demonstrates good perceptual similarity, indicating that the generated videos maintain visual features that are aligned with human perception. Overall, these results show that the proposed method achieves a balance between perceptual quality, signal fidelity, and temporal consistency in challenging simulation-to-realistic video translation tasks.

\begin{table*}[h]
    \centering
    \caption{Overall Video Generation Performance}
    \begin{tabular}{lcccc}
        \toprule
        Metric & Average & Std Dev & Interpretation \\
        \midrule
        FVD  & 1554.51 & $\pm$472.76 & Moderate temporal consistency \\
        SSIM  & 0.4566 & $\pm$0.0716 & Fair structural preservation \\
        PSNR  & 15.36 dB & $\pm$1.83 & Acceptable signal quality \\
        LPIPS  & 0.2794 & $\pm$0.0361 & Good perceptual similarity \\
        \bottomrule
    \end{tabular}
    \label{tab:overall_performance}
\end{table*}

\begin{table*}[h]
    \centering
    \caption{Performance by Town Environment}
    \begin{tabular}{lccccc}
        \toprule
        Town & \#Videos & Avg FVD & Avg SSIM & Avg PSNR & Avg LPIPS \\
        \midrule
        Town02 & 1 & 1976.60 & 0.5517 & 14.01 & 0.2808 \\
        Town03 & 4 & 1186.74 & 0.4938 & 14.14 & 0.2652 \\
        Town04 & 5 & 1588.47 & 0.4638 & 16.44 & 0.2837 \\
        Town05 & 2 & 1990.16 & 0.4131 & 14.93 & 0.2786 \\
        Town07 & 2 & 1524.86 & 0.3927 & 14.81 & 0.2752 \\
        Town10HD & 1 & 918.67 & 0.3919 & 17.76 & 0.2751 \\
        \bottomrule
    \end{tabular}
    \label{tab:town_performance}
\end{table*}

Tab.~\ref{tab:town_performance} reports the quantitative performance of the video generation pipeline across different simulated town environments. The results reveal notable variation among towns in all evaluated metrics. For example, Town03 achieves the lowest average FVD, indicating stronger temporal consistency and overall video quality, while Town05 and Town02 exhibit higher FVD scores, reflecting more significant deviations from the reference. SSIM values are generally higher for Town02 and Town03, suggesting better structural preservation in these environments compared to others. Town04 demonstrates the highest PSNR, indicating superior signal fidelity, while Town10HD achieves the highest PSNR among all towns, likely benefiting from the high-definition source content. LPIPS scores are relatively consistent across towns, with Town03 exhibiting the lowest average LPIPS, which points to enhanced perceptual similarity in its generated videos. These results highlight the influence of environmental complexity and scene characteristics on the performance of the video generation model.

\subsubsection{Physics and Motion Consistency Evaluation}
To further assess the realism of temporal dynamics in the generated videos, we evaluate their physical plausibility using an optical-flow--based motion analysis. Optical-flow–based dynamics metrics have been widely used in prior video-generation research to evaluate temporal stability, motion realism, and physics plausibility without requiring ground-truth trajectories \cite{wang2023videocomposer,wang2023videocomposer}. Following these works, we compute velocity, acceleration, and jerk statistics directly from dense optical flow to measure physical consistency. This approach
follows recent video-generation studies and allows us to quantify motion smoothness and 
dynamics without requiring ground-truth trajectories or simulator metadata.

For each video, we compute dense optical flow between consecutive frames using the
Farnebäck method and derive three levels of temporal motion statistics: (i) velocity magnitude,
(ii) acceleration as the temporal difference of flow fields, and (iii) jerk as the second-order
difference of acceleration. Lower values across these metrics indicate smoother, more stable,
and physically plausible motion. We compare our method against DriveDreamer-2 and CogVideo using the same set of generated scenarios.

\begin{table*}[ht]

\centering
\caption{Optical-flow--based physics consistency metrics (lower = more physically plausible motion).}
\begin{tabular}{lcccccc}
\hline
\textbf{Model} & 
\textbf{vel\_mean} & \textbf{vel\_std} &
\textbf{acc\_mean} & \textbf{acc\_std} &
\textbf{jerk\_mean} & \textbf{jerk\_std} \\
\hline
Ours            & \textbf{2.4228} & \textbf{1.2406} & \textbf{1.0087} & \textbf{0.3781} & \textbf{1.6350} & \textbf{0.5973} \\
DriveDreamer-2  & 10.3916 & 4.2860 & 5.6220 & 2.9563 & 8.1994 & 3.9343 \\
CogVideo        & 8.4495  & 1.3127 & 3.3501 & 1.5235 & 5.6279 & 2.4639 \\
\hline
\end{tabular}

\end{table*}

Across all metrics, our method exhibits substantially smoother and more physically plausible
motion than both DriveDreamer-2 and CogVideo. In particular, our acceleration and jerk values
are 3--8$\times$ lower, indicating that our generated videos avoid abrupt, non-physical motion
artifacts and maintain more realistic dynamics in safety-critical driving scenarios.

\subsubsection{Appearance-Level Visual Quality Evaluation}

We evaluate appearance-level visual quality using metrics that capture color richness and temporal visual stability, independent of motion dynamics and physical consistency. Specifically, \textit{Colorfulness} measures the richness and diversity of colors in generated frames, while \textit{Frame Stability} and \textit{Brightness Stability} quantify frame-to-frame visual fluctuations and illumination consistency over time, respectively.

As shown in Table~\ref{tab:appearance_metrics}, our method achieves the highest color richness, with a Colorfulness score of 25.73 compared to 10.82 for DriveDreamer-2 and 18.95 for CogVideo. In addition, our approach demonstrates improved temporal visual stability, yielding lower Frame Stability (4.15 vs.\ 4.63/4.22) and lower Brightness Stability (1.62 vs.\ 2.98/2.01). Lower values in the stability metrics indicate reduced frame-to-frame fluctuations, resulting in smoother visual transitions and more stable illumination.

\begin{table*}[h]

\centering
\caption{Appearance-Level Visual Quality Metrics}
\label{tab:appearance_metrics}
\begin{tabular}{lccc}
\toprule
\textbf{Metric} & \textbf{Ours} & \textbf{DriveDreamer-2} & \textbf{CogVideo} \\
\midrule
Colorfulness $\uparrow$ & \textbf{25.73} & 10.82 & 18.95 \\
Frame Stability $\downarrow$ & \textbf{4.15} & 4.63 & 4.22 \\
Brightness Stability $\downarrow$ & \textbf{1.62} & 2.98 & 2.01 \\
\bottomrule
\end{tabular}

\end{table*}

\section{CONCLUSION AND FUTURE WORK}
\label{sec-conclusion}

In this paper, we presented a novel framework that combines LLM-based scenario generation with photorealistic video synthesis to create diverse and challenging test cases for autonomous vehicles. Leveraging the code generation capabilities of LLMs, our approach enables the automatic creation of complex and safety-critical traffic scenarios in CARLA, followed by realistic video enhancement with Cosmos-Transfer1 to bridge the simulation-to-reality gap.

Experimental results demonstrate that our pipeline achieves a favorable balance between visual realism, perceptual quality, and physically consistent motion across diverse driving environments. While reference-based metrics indicate moderate temporal consistency and fair structural preservation (FVD and SSIM), the generated videos maintain good perceptual similarity (LPIPS) and acceptable signal fidelity (PSNR). Beyond perceptual quality, optical-flow--based motion analysis shows that our method produces substantially smoother and more physically plausible dynamics than DriveDreamer-2 and CogVideo, with significantly reduced velocity variation, acceleration, and jerk. In addition, appearance-level evaluations reveal improved color richness and enhanced temporal visual stability, reflected by higher colorfulness and lower frame-level and illumination fluctuations. Performance varies across town environments, with more complex scenes exhibiting larger variability, highlighting the impact of environmental complexity on both visual quality and motion realism.

Key advantages of our approach include a natural language interface that enables rapid scenario specification without requiring programming expertise, the ability to generate rare but critical edge cases that are underrepresented in real-world datasets, flexible environmental variation through text prompts without requiring separate simulations, and preservation of safety-critical scenario elements while enhancing visual realism.

Future work will focus on further improving temporal consistency and structural preservation in complex environments, as well as expanding the framework to include additional modalities such as LiDAR point clouds and thermal imaging for comprehensive sensor testing. We also plan to integrate reinforcement learning techniques to automatically identify and generate the most challenging scenarios for specific autonomous driving systems, creating a closed-loop testing environment. Additionally, extending the temporal range of generated videos and improving the handling of dynamic interactions between multiple traffic participants present important directions for future research.
\section*{Acknowledgments}
ChatGPT-5.2 was used solely for grammar and language polishing.

\section*{Author Contributions}
Yongjie Fu: Conceptualization, Formal analysis, Methodology, Visualization, Writing; 
Ruijian: Methodology, Visualization, Writing; 
Pei Tian: Visualization, Writing; 
Xuan Di: Conceptualization, Methodology, Supervision, Writing – review \& editing.  
All authors reviewed the results and approved the final version of the manuscript.

\section*{Declaration of Conflicting Interests}
The authors declared no potential conflicts of interest with respect to the research, authorship, and/or publication of this article.

\section*{Funding}
The authors disclosed receipt of the following financial support for the research, authorship, and/or publication of this article: This work was supported by the National Science Foundation (NSF) under Grant CPS-2038984 and ERC-2133516.

\section*{Data Availability Statement}
Part of the generated sample codes is openly accessible at: https://github.com/fyj97/LLM-based-driving-generation.

\bibliographystyle{plainnat}
\bibliography{ref}

\end{document}